# Leader-Follower 3D Formation for Underwater Robots

Di Ni[1], Hungtang Ko[1], and Radhika Nagpal[1]

[1] Department of Mechanical and Aerospace Engineering, Princeton University,
Princeton NJ 08544, USA
`{dn8856, hk1581, rn1627}@princeton.edu`

**Abstract.** The schooling behavior of fish is hypothesized to confer many survival benefits, including foraging success, safety from predators, and energy savings through hydrodynamic interactions when swimming in formation. Underwater robot collectives may be able to achieve similar benefits in future applications, e.g. using formation control to achieve efficient spatial sampling for environmental monitoring. Although many theoretical algorithms exist for multi-robot formation control, they have not been tested in the underwater domain due to the fundamental challenges in underwater communication. Here we introduce a leader-follower strategy for underwater formation control that allows us to realize complex 3D formations, using purely vision-based perception and a reactive control algorithm that is low computation. We use a physical platform, BlueSwarm, to demonstrate for the first time an experimental realization of inline, side-by-side, and staggered swimming 3D formations. More complex formations are studied in a physics-based simulator, providing new insights into the convergence and stability of formations given underwater inertial/drag conditions. Our findings lay the groundwork for future applications of underwater robot swarms in aquatic environments with minimal communication.

## 1   Introduction

Collectives of fish and fish-like robots have much to gain from moving in formation. In nature, thousands of fish schools coordinate their movement to migrate long distances, efficiently forage, and evade predators. Schooling is thought to confer many survival advantages, including hydrodynamic efficiency [1, 2]. By swimming in specific formations, it is hypothesized that fish leverage the fluidic interactions between their wakes to achieve higher thrust. Previous studies have documented many swimming formations adopted by schooling fish, including in-line swimming (one-behind-another), side-by-side (lateral) swimming, and staggered (diamond formation) swimming [3]. In the future, fish-inspired underwater robot "schools" can enable many important applications, from environmental monitoring of fragile ecosystems to search-and-rescue operations [4-6]. Underwater robot collectives can also benefit from moving in formation, for predictable spatial sampling, shared navigation, and potential energy savings. Fish-inspired robots can also serve as science platforms to test and validate hydrodynamic theories of how schooling formations in nature lead to energy benefits.

Formation control is an active area of research in robotics. Theoretical strategies for formation control have been quite diverse, ranging from reactive controllers such as the behavior-based methods [7] and potential field approaches [8], to more sophisticated controllers based on graph-theory [9, 35] and Lyapunov functions [10]. Some of these algorithms have been implemented on terrestrial robots [7, 11, 12], aerial robots [13-16], and unmanned surface vessels [17], focusing mainly on 2D formations and relying heavily on wireless communication and Global Positioning System (GPS). However, achieving underwater formation control poses significantly greater challenges. The primary difficulty stems from the high attenuation of radio frequency signals in underwater environments [4], which precludes the use of conventional methods employed by aerial or ground robots, such as GPS and Ultra-Wideband for navigation and coordination. Vision-based perception holds promise for enabling robot convoying in these challenging underwater settings. A supervised learning-based tracking approach has been implemented for two underwater robots, relying on vision to follow a target robot in the open sea [36]. However, this method requires extensive pre-annotated training data and has not demonstrated more complex formations beyond tracking. Overall, performing formation control with fully submerged underwater vehicles is still challenging.

The main contribution of this paper is the first experimental realization of multiple formations in three-dimensional underwater environments with an entirely vision-based leader-follower controller. The key innovations of our method include: (a) the absence of any explicit inter-agent communication with followers operating without advance knowledge of the leader's path, (b) the reliance on visual inputs exclusively, gathered by dual wide-angle cameras to ascertain the leader's bearing, pitch, heading angles, and the leader-follower distance, (c) the ability to achieve 3D formations in spite of perception errors and fluid effects. We use the BlueSwarm platform to demonstrate the successful implementation of fish-inspired formations, including in-line, side-by-side, and staggered swimming. We also demonstrate 3D formation, where robots follow a leader not just laterally but also vertically, using pitch angle control. Our hardware experiments focus on a single leader navigating a predefined trajectory—either a straight line or a circle—with one to two followers following on its side at different depths. Through simulation, we extend our results to include multiple followers, and we demonstrate how different formation positions have different stability properties underwater. Our research provides new insights, bringing us closer to future underwater robot swarms with enhanced visual perception and agility, that can move in complex 3D formations to achieve novel missions.

## 2    Related work

Formation control involves controlling robotic agents such that they move as a group while maintaining specific relative positions within the group, e.g. moving in a line or V shape or grid. There is a large body of work in theoretical control algorithms for formation control of general multi-robot systems [10, 17-21]. Common approaches include behavior-based strategies and artificial potentials, leader-follower strategies and graph-theory, and virtual structure strategies [10, 17, 18]. These strategies are not mutually exclusive and can be integrated or used complementarily.

In behavior-based formation [7, 22], robots respond based on a weighted combination of several desired behaviors, such as maintaining formation, moving towards a goal, and collision avoidance. The configuration of the desired formation is facilitated by the potential field, which enables automatic spatial distribution [8]. Behavioral strategies are characterized by their decentralized nature, minimal communication requirements, and simplicity of implementation. In leader-follower approaches, one or more agents are identified as leaders, and the other following agents keep a constant distance and orientation from the leaders [11, 23]. The leader-follower approach can be combined with graph theory to enable transitions between formations [9, 13]. In the virtual structure strategy, a centralized overseer controls the dynamics of virtual leaders and other agents in the formation align their motion with the virtual leaders [24]; a decentralized scheme was proposed in [25].

Theoretical formation control algorithms are not domain specific; robots are represented as idealized agents with strong assumptions about communication, unique identifiers, neighbor perception, well-specified and independent dynamics, etc. Implementing these algorithms on real robot systems has been challenging. Several examples consist of ground-based robots [7, 11, 12, 22], and above surface scenarios such as satellites [19] and unmanned aerial vehicles (UAVs) [13-16], mostly focused on 2D formations. Few attempts exist in underwater formation control, and are limited to a single horizontal plane [26, 27, 36]. In most of these implementations, the use of a wireless communication network and external global localization (GPS, or Vicon Motion Capture) are essential. Robots use the wireless network to exchange critical information, typically broadcasting their ID, pose, and velocity at high rates, allowing them to maintain specific positions relative to specific agents [4, 10, 13, 22, 37]. However, many challenges in swarm control algorithms stem from this dependence on communication. Underwater environments exacerbate these challenges due to the rapid attenuation of radio frequency signals in water, making wireless communication, external localization, and centralized control inaccessible. Furthermore, in nature fish schools use vision and local sensing to achieve formations, without explicit communication [1]. Therefore, there is a significant interest in developing algorithms for underwater swarm formation control, which rely on local perception rather than explicit communication [36].

## 3    Methodology

Our experiments are based on a 3D underwater swarm platform called BlueSwarm. This platform is comprised of 6 fully-autonomous miniature fish-inspired robots called Bluebots (Fig. 1b) with 3D maneuverability enabled by their multi-fin design and 3D perception enabled by fish-eye cameras and blue LEDs [28, 29]. Bluebots can control forward motion, turning, and depth independently. They are engineered for passive stability, allowing for yaw movements without the occurrence of roll or pitch. Each Bluebot is equipped with two wide-angle cameras and three blue LEDs for rapid detection of the position and heading of nearby robots with a small blind spot at the rear (~5 degrees). All experiments took place in two tanks: a small rectangular pool measuring 2.2 x 1.4 x 0.5 m³, and a large cylindrical steel tank with dimensions of 6.4 meters in diameter and 2.4 meters in depth (Fig. 1c). Note that both settings are very large compared to a Bluebot, with 1 meter being ~7.7 body lengths (BL). Overhead views of the robots were recorded using webcams positioned above each tank, the large tank also has a viewing window where a GoPro camera was placed to capture side-view videos. Once submerged underwater, Bluebots interact purely based on their onboard local computation and visual detection of neighbors. There is no communication possible with human or external computers.

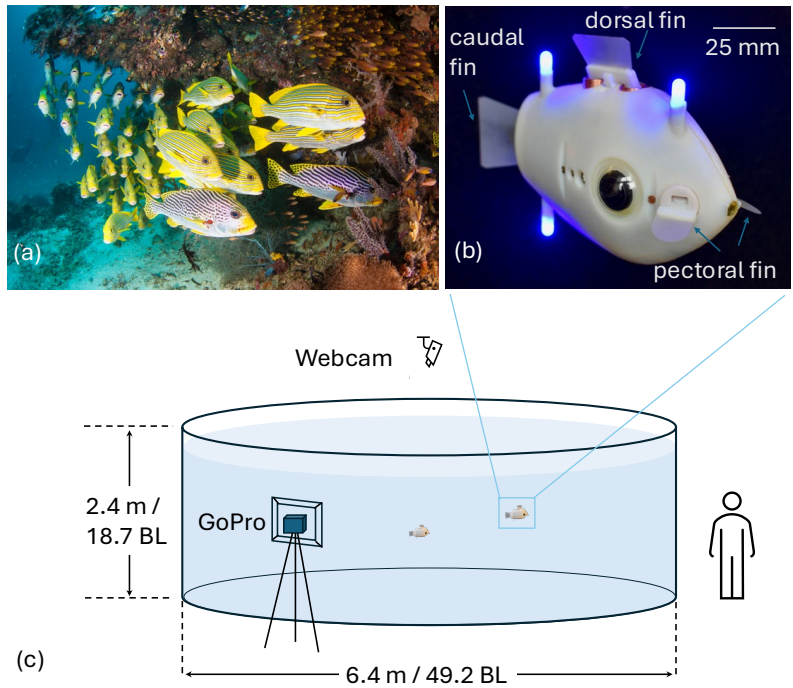

**Fig. 1.** We study fish schooling behaviors with the BlueSwarm robotic platform. (a) Fish exhibit collective swimming patterns in 3D (credit: istock). (b) Bluebot Robot: 3D motion is achieved by multiple fins enabling forward motion, turn in place, and depth control. The perception of leader robot (LEDs) is realized with onboard sensing and processing. (c) Tank facility showcased with a human figure to provide scale, equipped with both top and side cameras for recording.

This experimental underwater domain poses several key deviations from common theory assumptions: (1) *Perception errors:* Vision-based neighbor sensing underwater has many sources of error due to partial/full occlusion, surface reflections, and limited long-distance resolution. Furthermore, the heading error depends strongly on relative position due to the non-holonomic shape (e.g., robot pointed towards/away looks the same) (2) *Perception-Control Loop:* Bluebots have a limited vision-processing rate of 5 Hz, which prevents more sophisticated forms of neighbor estimation and prediction. Even with higher computational resources, online natural vision processing remains slow in both underwater and aerial vehicles. (3) *Underwater dynamics:* Fluid forces significantly impact motion controllability but remain difficult to model or sense. Bluebots are never stationary and drift constantly even when all fins are off. They have relatively large inertia compared to fluid drag and cannot brake or swim backward. In the context of formation control, robot's trajectories are also being impacted by neighbors' wakes in ways that are still barely understood. Despite the major advances in Computational Fluid Dynamics (CFD), it is still intractable to simulate the complex fluid interactions for free-swimming Bluebot formation control. *Our goal therefore is to investigate experimentally whether formation control can be achieved that is robust to the constraints and limitations of underwater perception and control, and to interrogate the gap between theory models and experimental realizations.*

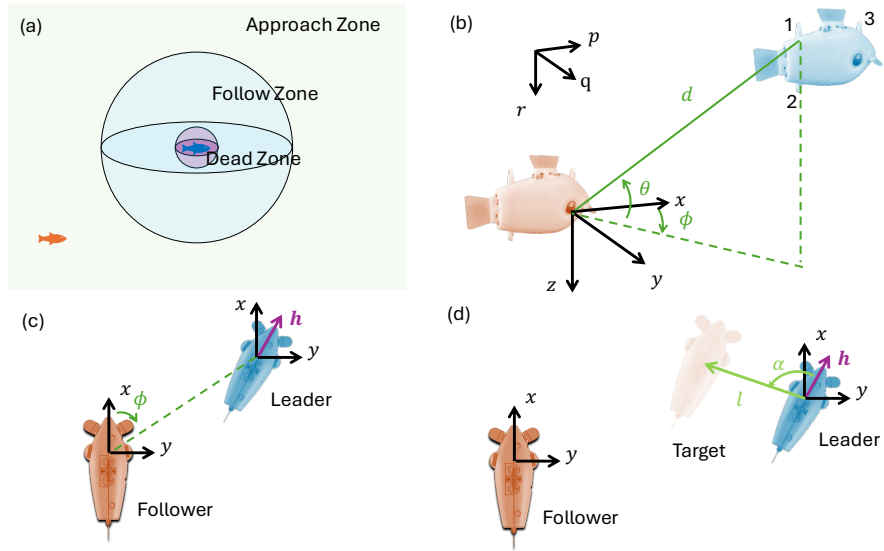

**Fig. 2.** (a) Defined zones as seen from the perspective of the follower (orange fish). (b) From the follower's point of view, the bearing $\phi$, pitch $\theta$, and distance $d$ to leader. (c) The leader's heading direction $\boldsymbol{h}$. (d) A target position defined by rotating the heading vector with an angle of $\alpha$ and shift with a distance of $l$.

### 3.1 Formation Control Algorithm and Vision Processing

We chose an algorithm that combines behavior-based [7] and leader-follower methods [9, 23], and extend the methods to 3D space. We chose this reactive approach because of its simplicity of assumptions and potential for robustness to sensing error and leader trajectory changes. The system defines two agent roles: *the leader (L),* tasked with navigating a predetermined path, and *the followers (F),* who lack any prior knowledge of the leader's trajectory, but are charged maintaining a position relative to the leader. The

6leader swims in a straight line or a circle with open loop control at a predefined depth underwater. The follower achieves formation in two steps: a perception phase, where the leader's pose and heading are ascertained, followed by a movement phase, where the actuation of fins is determined by the follower's position relative to the leader.

The follower robots utilize vision exclusively to acquire critical information about the leader. The leader Bluebot has its blue LEDs on. If there are multiple followers, then those Bluebots will have their LEDs off. Future variations can have two colors of LEDs, to distinguish leaders from followers. The vision processing by the follower has several steps:

(i) *Blob detection and LED parsing:* In each processing cycle, lasting 0.2 seconds, the follower robot captures a pair of images from its left and right cameras. It then detects the leader's characteristic features (*pqr* coordinates) of three blobs using the custom-designed blob detection algorithm introduced in our prior work [29, 30]. The vision algorithm processes the input images and discerns the sequence of the three LEDs on the leader, where the order of detected LEDs is crucial (Fig. 2b). As the robot prohibits pitch movement, the two posterior LEDs 1 & 2 remain stacked. To avoid miscounting water surface reflection as LEDs, the lowest two blobs are selected to calculate distance $d$ using geometry mapping, similar to our prior work [30].

(ii) *Leader Pose Determination:* Bearing can be calculated from LEDs 1 &2 using

$$\phi = \arctan\left(\frac{q}{p}\right) \tag{1}$$

where p and q represent the coordinates of the leader's blobs in the pqr coordinate system from the follower's perspective in camera space. To determine the direction $\boldsymbol{h}$ in which the leader is heading (Fig. 2c), the follower must use the 3rd LED as a reference. A blob that closely matches the pitch of the two previously identified blobs is recognized as the third LED. With the Bluebot setup, we establish a threshold for pitch difference at 6 degrees, based on experimental data as the maximum pitch variation at which a robot can detect its neighbor's LED. Blobs not meeting this criterion are considered reflections and thus ignored.

(iii) *Target Pose (Fig. 2d):* To maintain a position alongside the leader, the follower computes the new target pose using:

$$\boldsymbol{x}_{goal} = \boldsymbol{x}_{leader} + l\,\boldsymbol{h} \cdot \boldsymbol{R}_z(\alpha) \tag{2}$$

where $\boldsymbol{x}_{leader}$ is the leader's pose vector in $xyz$ coordinates, which is translated from the camera's representation to real physical world dimensions. The scalar $l$ denotes the predetermined following distance in the xy plane between agents. The vector $\boldsymbol{h}$ refers to the leader's heading vector, and $\boldsymbol{R}_z(\alpha)$ is a 3 × 3 rotation matrix along $z$ direction with a predefined following angle $\alpha$.

(iv) *Target Pitch (Fig. 2b):* The depth control of the follower is controlled separately based on the pitch angle of the leader observed from cameras:

$$\theta = \arctan\left(\frac{r}{\sqrt{p^2 + q^2}}\right) \tag{3}$$

A positive angle means the leader is lower than the follower and vice versa. The follower switches the dorsal fin on to move down, and switches the dorsal fin off to move up using positive buoyancy.

The follower's movement phase is governed by an enhanced zone concept, originally introduced in [7] and now extended into 3D space. As shown in Fig. 2a, the space is divided into three zones: the approach zone, the follow zone, and the dead zone. In the *approach zone*, the follower is at a considerable distance from the leader. It designates the leader's position as the target and moves toward it at the maximum speed. Upon nearing the leader, the follower transitions to the *follow zone* and aims for a specific formation position defined by a desired angle $\alpha$ and a desired lateral distance of $l$ from the leader. The speed of the follower in the follow zone is set to be linearly dependent on the distance between L and F, which varies from a maximum at the farthest edge of the follow zone to a minimum at the inner edge. The depth of the robot is controlled by adjusting the dorsal fin to follow the leader at a desired pitch angle $\theta_0$. When the follower gets too close to the leader, it enters the *dead zone*, and all fins except the dorsal fin stop actuation to avoid collisions.

## 4 Experimental Demonstrations and Results

Using the setup and algorithms outlined above, we demonstrated and tested three formations that are commonly observed in fish schools: *in-line* (follower behind the leader), *side-by-side* (follower is to left or right of the leader, at same or different depths), and *staggered* (two followers). In each case, the follower is attempting to maintain a pre-specified distance, bearing, and depth relative to the leader. We include videos of all experiments here (https://youtu.be/9WoqaPVQCfU). We use tracked trajectory data to determine the success of the formation control. We processed the overhead camera recordings using a custom MATLAB program to track the horizontal movement of the robots, and the vertical displacement of the robots was retrieved using their onboard depth sensor. We conducted two sets of experiments, 1 leader + 1 follower in the large tank and 1 leader + 2 followers in the small pool. In the simulation section, we extend our results to larger numbers of followers.

### 4.1 Follow a leader swimming in a straight line in the same plane

In the first experiment, we demonstrated a follower robot's capability to track a leader either directly in-line behind or side-by-side, maintaining alignment within the same horizontal plane. The leader was programmed to swim in a straight line. The follower was configured with an approach threshold of 500 mm. When the distance was larger than this threshold, the follower targets following behind the leader in in-line position. If the distance fell below this threshold, the follower would adjust to position itself beside the leader at a 90° angle and a lateral distance of 200 mm (1.5 BL). The pitch angle was assigned to remain within a range of [-1, 1] degrees so that the follower maintained the same planar alignment as the leader.

Tracking data shows that the formation was successful. Fig. 3a shows the initial states of the robots, positioned at a distance greater than the approach threshold, each with a random heading orientation. Fig. 3b shows the trajectories from start to 15s when the follower targeted to follow right behind the leader. Once the follower entered the follow zone, it adjusted its target position to the leader's left, maintaining a side-by-side formation (Fig. 3c and 3d). Figure 3e shows that the follower kept about 250 mm from the leader in the follow zone.

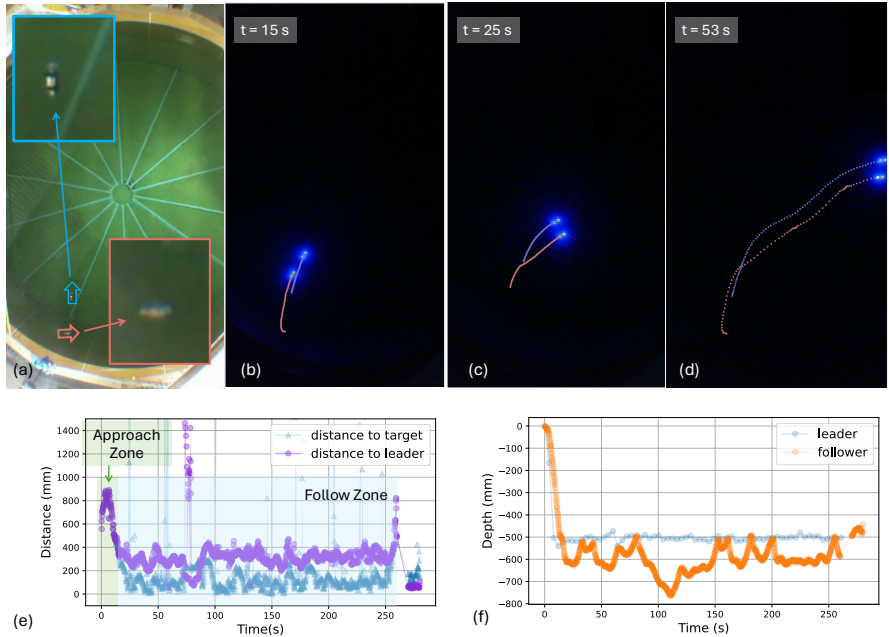

**Fig. 3.** (a) Robots in the big tank with random initial positions, showing the leader robot encircled by a blue arrow and the follower robot by an orange arrow. (b) The follower robot (red) trails directly behind the leader (blue) in the approach zone. (c) Upon entering the follow zone, the follower robot positions itself beside the leader. (d) Trajectories of the leader and follower after 53 seconds of movement. (e) Distances to the leader and the targeted pose as observed by the follower robot. (f) Vertical positions of leader and follower robots from the onboard depth sensor.

We also see some spikes in the data that indicate specific events and noises. A collision at 73 seconds distorted vision and distance measurements until 78.7 seconds. At 260s the leader turned off its LEDs and laboratory lights were switched on at 270s. Post 270 s, the vision algorithm was deceived by ambient lights, erroneously detecting the leader at a much closer range than actual. There are noises in calculating target distance (equation 3), which shows up as sporadic spikes (1.38%).

The follower was able to maintain the formation despite these visual errors. We observed an increase in the vision processing error rate upon integrating and analyzing the third LED to determine leader heading. Prior work with Bluebots used only two LEDs to determine distance from neighbors [29]. The addition of a third LED is essential for formations because a follower needs leader heading information to ascertain its target position, e.g. left or right of the leader. The error in heading determination depends on the relative orientation of the leader and follower, e.g. LEDs 1 and 3 may appear as a single blob when the leader is directly in front at [-30, 30] degree angle.

Reflections from the water surface are more likely to be erroneously identified as a third LED in such cases. By implementing additional error corrections (discussed in section 3) the follower managed to maintain an accurate estimate of the leader's pose and heading. However, it did not eliminate the occurrence of incorrect estimations.

Figure 3f shows the follower's adjustments to match the depth of the leader. Initially, from 0 to 14 seconds, the follower successfully descended to align with the leader's depth plane. Subsequently, from 14 to 73 seconds, the follower exhibited oscillations around the leader's depth, with a maximum deviation of approximately 140 mm, equating to 1.06 BL. Upon colliding with the wall at 73, the follower's path showed increased deviation from the leader's depth (73s – 110s), peaking at a maximum divergence of 264 mm (2.03 BL). The follower gradually recovered, re-establishing a following pattern with an approximate deviation of 1 BL.

### 4.2    Follow a leader swimming in a circle at different depths

To further assess the algorithm's robustness, we conducted tests with the leader performing a circular swim and the follower navigating alongside at various depths. In the follow zone, the follower's target position was defined to be at a ±90° angle relative to the leader's head, prompting it to either widen or tighten its orbit in accordance with the leader's trajectory. The follower was programmed to maintain a lateral gap of 150 mm and a pitch angle between [-40, -45] degrees with respect to the leader, resulting in a desired Euclidean distance of 196 – 212 mm.

Fig. 4 shows the performance of following on leader's outside and inside. The side view images validate the follower's ability to maintain a deeper position in the water compared to the leader. In scenarios where the follower is required to follow in a larger circle, it must swim at a faster rate to keep pace with the leader's side. The top view confirms the follower's capacity to preserve its relative position to the leader. However, due to the maximum speed constraints of the Bluebot, the follower ended up adopting a lagged angle, resulting in a staggered formation. In contrast, in the case of following within a smaller circle, the follower tended to overshoot during the leader's turns, leading to an erratic following path.

Figure 4c plots the follower's estimated distances to the leader and the target, highlighting a pattern of greater separation when following a larger circle compared to a smaller one. In the larger circle following case, the distance to the leader averaged around 350 mm, which was ~1 BL larger than the intended following distance, likely due to insufficient response speed. In contrast, the average following distance in the smaller circle was approximately 0.5 BL less than the target, suggesting a tendency for overshooting. The smaller circle's path also demonstrated more discontinuities, attributable to the visual processing challenges presented by the necessity of executing tight turns to maintain clear sight of the leader. The depth profiles in Fig. 4d show that the follower was able to maintain to follow at deeper depths around 700 mm depth, though with oscillation, which might be caused by the large following pitch range of 5 degrees assigned to the follower.

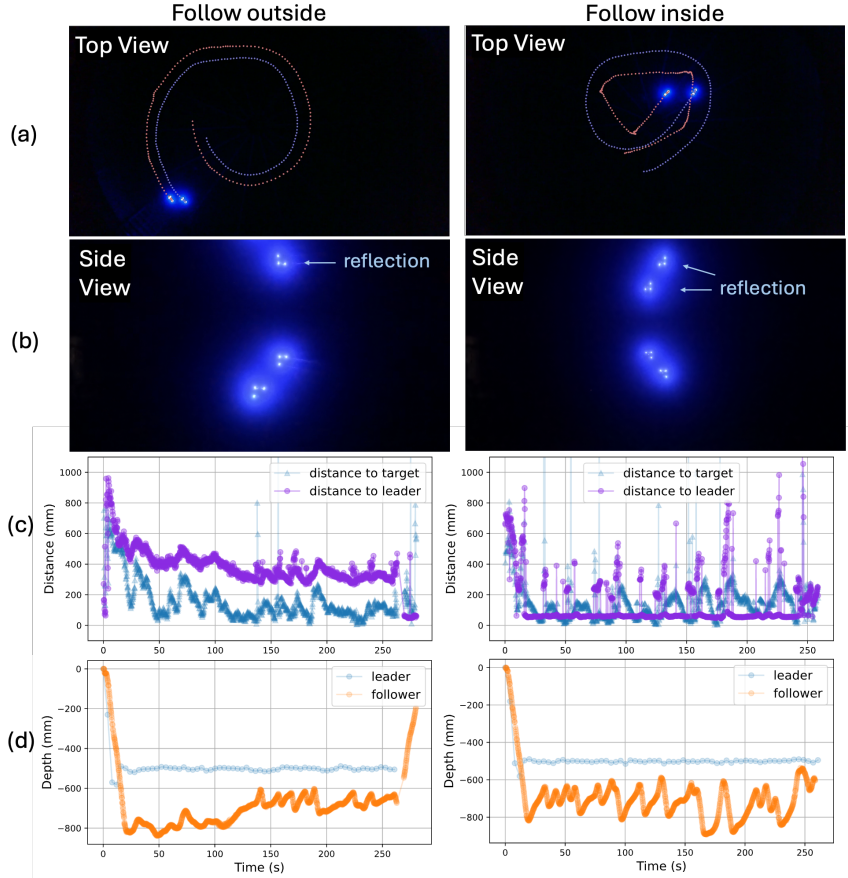

**Fig. 4.** A leader robot is programmed to swim in a circle in open loop control, and a follower robot follows in a larger circle and a small circle. (a) The trajectories for leader (blue) and follower (red). (b) Side views of the following performance. (c) The estimated distances to the leader robot and to the target position from the follower's perspective. (d) Diving depths of both robots from onboard depth sensors.

From our experiments, it can be recognized that not all formations pose the same level of challenge for the robot's visual perception and maneuverability. Following behind a straight-swimming leader is comparatively simple. Once the follower aligns with the targeted position, maintaining a straight trajectory involves minimal turning, which is the simplest form of navigation. Here, the follower primarily propels forward, a direction where it exhibits maximum control and maneuverability. On the other hand, tailing a leader that engages in continuous turning significantly increases the difficulty. The robots must constantly adjust to provide the appropriate centrifugal force to turn, all while keeping up with the leader's speed and maintaining the correct relative positioning. Moreover, following on the outside versus the inside brings its own set of challenges. When following externally, the follower must increase its speed to compensate for a greater turning radius. Conversely, an internal following demands rapid adaptation to the leader's rapid changes in position and heading. This requires the robot to have not only better maneuverability during turns but also a faster rate of vision processing.

### 4.3 Two followers with a single leader moving in a line.

In Fig. 5, we expanded our experiments to include one leader and two followers on both sides of the leader within the small pool, demonstrating the algorithm's applicability to multi-agents scenarios. These trials were confined to a smaller pool due to facility constraints; although the overhead camera for the larger tank was positioned to capture the entire area, its resolution was insufficient to identify follower robots when their LEDs were turned off. As shown in the video and figures, the robot initially started in the wrong position by then quickly achieve the correct side by side formation. They are able to follow the leader even when it makes an unexpected turn.

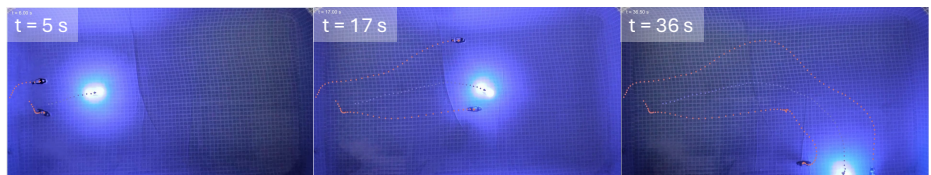

**Fig. 5.** Two followers following a leader's sides. Each frame captures a different timestamp, denoted by t = 0s through t = 36s, demonstrating the followers' ability to maintain formation over time (trajectory is marked by a dotted trail).

## 5 Simulation Demonstrations and Extensions

To further evaluate our formation control algorithm, we conducted simulations using a BlueBot simulator, previously validated and described in [30]. The simulator replicates the Bluebot's hydrodynamic design, where agent movement is generated by the actuation of pectoral, dorsal, and caudal fins. External forces in the simulations are based on physical principles, including a thrust proportional to fin flapping frequencies, and a fluid drag proportional to the velocity squared. However, it does not model complex fluid effects, such the wake of neighboring robot. In line with the sensory capabilities of actual Bluebots, simulated agents detect each other by processing the LED blobs from neighboring robots. Agents have a blind spot and blobs can be obstructed by neighbors. However, the simulated vision is not subject to other parsing errors and have a larger effective range compared to experiments. The simulator provides a reasonable approximation to the real hardware system that allows us to systematically investigate more initial conditions, more robots, and test potential extensions to the algorithms.

### 5.1 From Zones to Hyperbolic Control

As a first simulation study, we replicate our experimental formation study in Section 4 and test the robustness to random initial conditions. Fig. 6 shows that an agent can successfully follow on both the inside and the outside of the leader regardless of its initial position. In both cases, the follower successfully estimated the relative location and heading of the leader and eventually converged to its goal location, i.e. 200 mm away from the leader laterally. Followers instructed to follow on the right and on the left of the leader moved in circles of different radii without building any model of the leader's behavior. We verified that this behavior over 30 initial conditions.

However we noticed that the follower did not settle down to a perfectly circular trajectory (Fig. 6 a,d), which we also observed in the experiment. Instead it fluctuated around the targeted position (Fig. 6, b,e) and distance (Fig. 6 c,f). This is because Bluebots have a relatively large inertia compared to the fluid drag (Re~$O(10^4)$ ) [31], and have limited ability to brake actively, both features captured in our simulation. As a result, simulated agents could not be controlled perfectly and often drifted past their target. Critically, following on the outside versus inside required different control inputs. Following on the right side of the leader requires a larger turning radius and a higher speed, calling for a higher frequency from the caudal fin (Fig. 2b) compared to the pectoral fin. The opposite is true when following on the inside. The zonal approach was not able to automatically adjust for such a difference.

To further increase the accuracy of formation control, we investigated a variant of the control approach where the zonal separation (Fig. 2a) was replaced with a hyperbolic tangent function. In this modified approach, the driving frequency of the caudal fin (and hence the thrust) increased smoothly with distance between the following agent and the target position. This approach allows the follower to automatically adapt to the turning radii required to maintain the formation. When following on the outside, the follower is always farther from the target position compared to following on the inside, resulting in a higher caudal fin frequency and thrust. Figure 6 (g,h) shows that the modified approach indeed led to more accurate formation control. Compared to Fig. 6 (c,f), the following agent stayed closer to the target position for both cases. In clear contrast with the zonal approach, Fig. 6 (g,h) also shows that the distance between agents converged better to the set distance at 200 mm. In practice, current Bluebots have limited smoothness in their speed control, however future versions can use such functions to achieve even higher accuracy and smoothness.

### 5.2 Complex Hexagonal Formation

We also used the simulator to test more complex formations with multiple agents. In particular, we wanted to investigate whether certain formation positions were easier or harder than others. Formation control algorithms typically ignore this aspect, assuming all positions are the same. However non-holonomic vision and motion, plus the impact of inertial motion in water suggest otherwise. In our experiments we observed that in-line formation (follower directly behind) makes it difficult to detect leader heading, while 3D staggering improves behavior by reducing collision avoidance.

Fig. 7 (a-b) shows an example of six agents following in a hexagon pattern around a leader. Such a formation around a circular leader leads to multiple circular trajectories of different radii (Fig. 7a). While most agents fell into their prescribed locations with respect to the leader (Fig. 7b), the agents right behind (#3) and in front of (#6) the leader failed to maintain its position as well as the other followers. Agent #3 was limited by its ability to calculate the leader's heading. Situated directly behind the leader, it could not see all 3 LEDs of the leader (Fig. 2b). Limited visibility is also partially responsible for agent #6's imperfect tracking. Bluebots had a blind spot right behind them, which was captured in our simulator. Therefore, the agent in front of the leader often lost sight of the leader temporarily before finding it again when it was at a bearing angle behind. Furthermore, agent #6 needed to constantly readjust its pose by braking and turning at the same time to maintain its position, which was difficult for Bluebots.

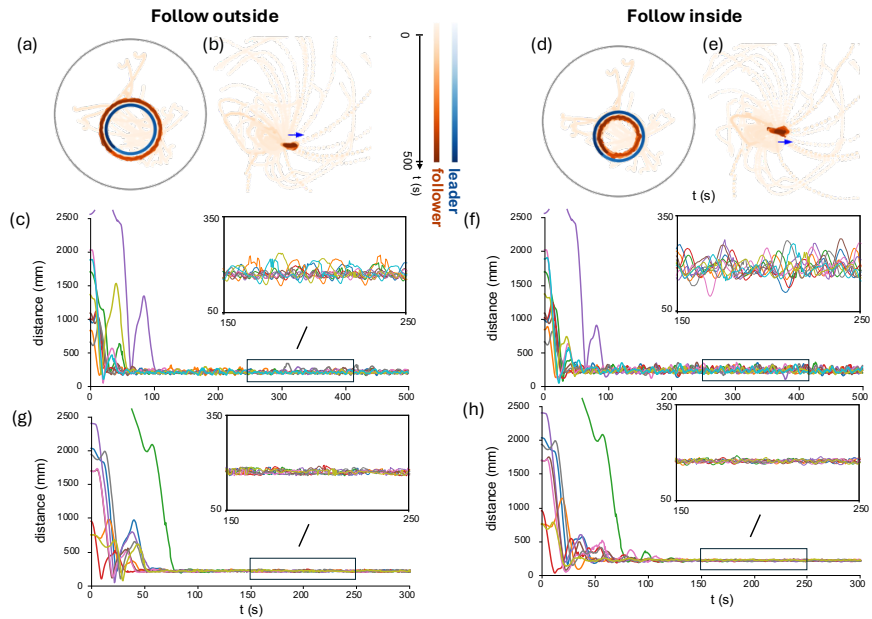

**Fig. 6.** (a, d) Trajectories of the leader (blue) and the follower (red) in the lab frame. The darkness indicates time and 30 numerical trails starting from different initial locations are shown. (b, e) Trajectories of the following agent with respect to the leader. Blue arrow indicates orientation of the leader. (c-h) Time series of the distance between leader and follower. Different colors show different numerical trials. The zonal approach was used in (a-f), and the modified algorithm used in (g,h).

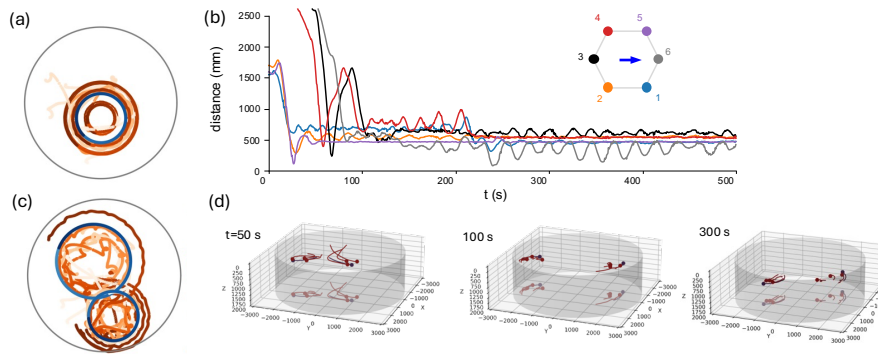

**Fig. 7.** (a) Trajectories of six agents forming a hexagon pattern around a leader moving in a circle. (b) Distance between the agents and the leader. Different colors represent different followers. (c - d ) Trajectories of multiple followers. One leader (top) moves counterclockwise and the other (bottom) moves clockwise.

Fig. 7 (c-d) demonstrates that our algorithm can be applied to scenarios with more than one leader. As the two leaders circled in opposite directions with different radii, the agents followed the leader the closest to them. The leaders occasionally crossed paths and the followers reshuffled. The relationships between leaders and followers emerged automatically from our algorithm. This could be a desired feature for applications where several leaders move along preprogrammed trajectories or are remotely controlled while multiple followers remain on their sides based on proximity.

# 6     Conclusion and Future Work

The main contribution of this paper is the experimental realization of an entirely vision-based 3D formation control algorithm, using a behavior-based leader-follower strategy. Unlike previous work in ground and aerial robots, no explicit communication or external global localization is used. We demonstrate the successful implementation of fish-inspired formation behaviors, such as in-line and side-by-side swimming, using the BlueSwarm robotic platform. We also demonstrate 3D formations, where robots follow a leader not just laterally but also vertically, using pitch angle control. These formations are important for scientific studies of hydrodynamic hypotheses, to understand how energy savings arise from the fluid interactions of vortex shedding in fish pairs, and how to extend those energetic savings to future underwater robots. Furthermore, underwater robots moving in formation can leverage better algorithms for spatial sampling and group navigation in complex environments. Our research also provides new insights into the theory2robots gap for underwater systems; we find that reactive algorithms are able to successfully achieve formations in spite of perception errors and limited perception rate, but that inertial drift and non-holonomic control makes certain formation positions harder than others. Overall, our work brings us closer to realizing real underwater robot swarms, that can achieve more complex missions such as energy-efficient long-distance navigation.

An important area of future work is improving physical underwater swarm platforms, enhancing both visual perception and agility. Blue LEDs are limited to lab settings, and as computation becomes smaller and cheaper, underwater robots can include more natural vision processing that will allow them to swim together in cluttered environments such as coral reefs or underwater wrecks [36, 37]. Bluebot multi-fin actuation provides substantial 3D agility, but only at slow speeds (1 BL/s) and with low thrust. Future fish-inspired robot platforms can take advantage of higher speed/thrust flapping actuation that can operate in natural environments [32-34]. However, these platforms must also include agility in yaw and depth maneuvers in order to do 3D leader-follower behaviors. Such platforms could eventually even follow fish or accompany fish schools, providing us with unprecedented insights into these amazing natural systems.

**Acknowledgments:** This work was supported by ONR N00014-21-S-F003, ONR N00014-22-1-2222, and the James McDonnell Fellowship. Inspired by a recent initiative to increase awareness and mitigate citation bias [38], we include a gender citation diversity statement using manually compiled data. Our references contain 2.7% woman (first author)/woman (last author), 24.32% man/woman, 5.41% woman/man, 64.86% man/man, 2.7% non-identified.